\documentclass[letterpaper, 10 pt, conference]{ieeeconf}
\IEEEoverridecommandlockouts
\usepackage{cite}
\usepackage{amsmath,amssymb,amsfonts}
\usepackage{graphicx}
\usepackage{textcomp}
\usepackage{xcolor}
\usepackage{multirow}
\usepackage{subcaption}
\def\BibTeX{{\rm B\kern-.05em{\sc i\kern-.025em b}\kern-.08em
    T\kern-.1667em\lower.7ex\hbox{E}\kern-.125emX}}

\usepackage{tikz}
\def\checkmark{\tikz\fill[scale=0.25](0,.35) -- (.25,0) -- (1,.7) -- (.25,.15) -- cycle;} 

\makeatletter
\newcommand\notsotiny{\@setfontsize\notsotiny\@vipt\@viipt}
\makeatother

\usepackage{lettrine}
\usepackage{pdfpages}
\usepackage{hyperref}

\hypersetup{
    colorlinks=true,
    linkcolor=black,
    filecolor=magenta,      
    urlcolor=cyan,
    citecolor= black,
    }

\title{\LARGE \bf Robotic Packaging Optimization with Reinforcement Learning\\
}

\author{Eveline Drijver$^{1}$, Rodrigo Pérez-Dattari$^{1}$, Jens Kober$^{1}$, Cosimo Della Santina$^{1}$ and Zlatan Ajanović$^{1}$
\thanks{$^{1}$Authors are with Cognitive Robotics Department,
        Delft University of Technology, 2628 CD Delft, The Netherlands
        {\tt\small E.A.Drijver@gmail.com, \{R.J.PerezDattari, J.Kober, C.DellaSantina, Z.Ajanovic\}@tudelft.nl}}%
 }

\begin{document}

\maketitle
\thispagestyle{empty}
\pagestyle{empty}

\begin{abstract} 
Intelligent manufacturing is becoming increasingly important due to the growing demand for maximizing productivity and flexibility while minimizing waste and lead times. This work investigates automated secondary robotic food packaging solutions that transfer food products from the conveyor belt into containers. A major problem in these solutions is varying product supply which can cause drastic productivity drops. Conventional rule-based approaches, used to address this issue, are often inadequate, leading to violation of the industry's requirements. Reinforcement learning, on the other hand, has the potential of solving this problem by learning responsive and predictive policy, based on experience. However, it is challenging to utilize it in highly complex control schemes. In this paper, we propose a reinforcement learning framework, designed to optimize the conveyor belt speed while minimizing interference with the rest of the control system. When tested on real-world data, the framework exceeds the performance requirements (99.8\% packed products) and maintains quality (100\% filled boxes). Compared to the existing solution, our proposed framework improves productivity, has smoother control, and reduces computation time. 
\end{abstract}


\section{Introduction} 
\label{sec:intro}
%
%
%
With the manufacturing industry aiming towards flexibility, productivity, quality, and mass customization \cite{b41}, the adoption of intelligent manufacturing becomes increasingly important. This is especially relevant for highly competitive fields such as the food packaging industry \cite{b46}. One of the primary challenges in the food packaging industry is to enhance productivity while simultaneously reducing lead times. Achieving this objective entails developing standardized packaging solutions that minimize waste while maximizing food production rates.

\begin{figure}[t]
\centerline{\includegraphics[width=1\linewidth]{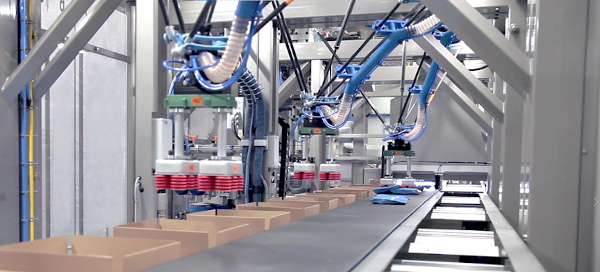}}
\caption{An industrial secondary packaging solution where robots coordinate to pick and place products into packaging boxes.}
\label{fig:spider}
\end{figure}

%
%
%
%
Packaging solutions can be divided into two parts: primary and secondary. Primary packaging solutions are designed to handle naked and unpackaged products, such as cookies or chocolates. In contrast, secondary packaging solutions are designed to handle already packaged products that require placement into a secondary container, e.g., a box (see Fig. \ref{fig:spider}). As a consequence, secondary packaging solutions should be able to handle uncertainties generated by the upstream production process, such as variance in the product supply rate.

Commonly, the problem of varying product supply is addressed using rule-based engineered solutions. Nevertheless, depending on the task at hand, these solutions may fall short when aiming to satisfy specific task requirements. For instance, in some scenarios it is critical that boxes do not leave the machine unfilled, and the speed of the conveyor belt that transports the packaging boxes must be controlled to comply with this requirement. Unfortunately, this problem remains challenging for rule-based methods, since it requires multiple robots to coordinate when packaging the products, and handling all the effects that a change in the speed of the conveyor belt generates is not intuitive. Consequently, these methods often compromise the performance of the machine, especially in edge cases, commonly violating requirements regarding productivity.

In this context, data-driven methods become appealing, since they are able to provide robust solutions without requiring an understanding of all of the specific details of the system. More specifically, Reinforcement Learning (RL) allows for finding solutions by trial and error, addressing even the most challenging edge cases with sufficient training. This is particularly valuable when the aim is to surpass the performance of existing suboptimal rule-based engineered solutions. Moreover, the same framework can be employed to obtain controllers for different machines and scenarios, without requiring a re-engineering process, as long as the problem’s objective remains constant. Nevertheless, although RL methods have demonstrated success in various simulated environments, their adoption in real-world applications has been relatively slow. A significant factor contributing to this is the limited focus of RL research on practical industry applications, resulting in a substantial disparity between the design of current experimental RL setups and the often ill-defined nature of real-world systems, which require addressing issues like system delays, safety and performance constraints \cite{b8}. Furthermore, in industrial scenarios, RL may be utilized to tackle only a subset of a larger problem, necessitating its coexistence with other control systems while minimizing interference with them \cite{b54}. 

In this work, we aim to investigate the feasibility of RL in the industry for robotic packaging optimization. We propose an RL approach designed to optimize the box conveyor belt speed in order to maximize the performance of the robotic packaging machine under varying product supply. With this method, predictive behavior for varying product supply is learned while satisfying the machine's performance constraints. The design encourages smooth control of the conveyor belt and handles a limited availability of real-world product supply data properly. For practical reasons, the method is not validated on a physical machine. Instead, we employ real-world data (that can be replayed on a simulator) to validate it, where it obtains a higher performance when compared to a rule-based method currently used in the industry.

Our contributions are:
\begin{itemize}
    \item a method that uses planned delay to enable integration of RL solution in a highly complex control scheme, with minimal interference with other controllers.
    \item a RL framework capable of learning robust behaviors from limited real-world data.
    \item a RL solution based on the presented method and framework for optimizing real robotic solutions in the food packaging industry.
\end{itemize}


\section{Related Work} 
Existing research on RL deployment in real-world settings has identified key challenges, such as dealing with system delays, partial observability, policy inference, and satisfying safety and performance constraints \cite{b8}. However, current approaches only address a subset of these challenges, hindering the fast adoption of RL in the real world. While successful applications of RL in the industry include Google DeepMind's data-center projects \cite{b44}, the robotics packaging industry lacks similar success stories. Previous research, such as \cite{b43}, has explored the feasibility of RL in real-world settings using physical industrial robotic manipulators. In this study, we aim to extend RL adoption by evaluating the feasibility of RL in a real-world robotic problem within an industrial setting, rather than focusing on physical implementation in an experimental setup.

Safe RL is a popular research field concerned with ensuring system performance and safety constraints during learning and deployment \cite{b4, b42}. This is particularly important for the transfer of RL towards industry, where strict safety and performance requirements must be met. Various methods have been used to address these requirements. For instance, \cite{b3} is devoted to model-free RL for policy learning using constrained Markov decision processes. Additionally, Safety Layers \cite{b36} can be used to safely explore Markov decision processes, while Risk-Averse Robust Adversarial RL \cite{b38} is useful for learning robust risk-aware probabilistic models. Another approach is Bayesian Controller Fusion \cite{b39}, which allows for accelerated safe learning with suboptimal controllers. However, most research is limited to simulated or simplistic real-world tasks, without formal guarantees for hard or probabilistic constraints. In contrast, our work aims to test the feasibility of RL in a complex real-world robotic problem with strict performance constraints.

\section{Problem Statement} 
\label{sec:problemstatement}
First, we introduce the robotic packaging process that we are optimizing, followed by the specific belt speed optimization problem that we solve using our RL framework.

\subsection{Robotic Packaging Process} 
\label{subsec: robotic packaging machine}

\begin{figure}[t]
\centerline{\includegraphics[width=1.\linewidth]{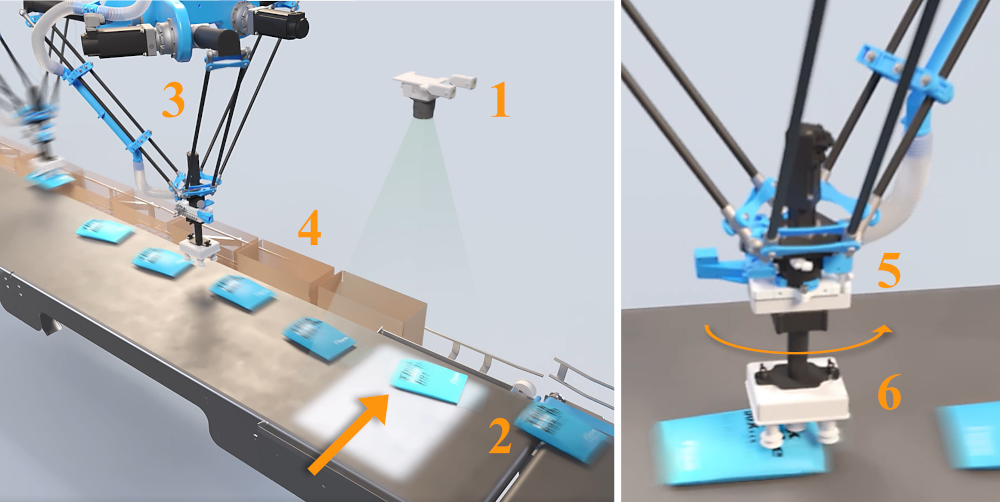}}
\caption{Components of the robotic packaging process.}
\label{fig:OverviewMachine_option2}
\end{figure}

A simplified view of the machine's internal operation is shown in Fig. \ref{fig:OverviewMachine_option2}, where a vision system (1) detects the class and location of pre-packaged food products (2) on the conveyor belt. The packaging process involves multiple delta robots (3). These robots pick up the packaged products and place them in a container (4), e.g., boxes or pockets. The right subfigure depicts a close-up of a delta robot with a rotational end-effector (5) and vacuum gripper (6). 
The packaging solution optimizes the pick-and-place sequence to maximize the throughput of products per minute.

During deployment, the machine needs to satisfy the following requirements: 
\begin{itemize}
    \item At least 99.8\% of the supplied products must be placed in a box.
    \item No empty boxes may leave the machine. 
    \item No partly filled boxes may leave the machine.
\end{itemize}

To test the effectiveness of the packaging machine with respect to these requirements, we use a simulator of the machine that simulates realistically system dynamics, delays and product infeed. For measuring the overall effectiveness of the machine we use the Overall Equipment Effectiveness (OEE) industry standard \cite{b47}, which is widely used in the manufacturing industry. The OEE metric determines the effectiveness of the machine as a ratio between actual and theoretical output \cite{b47} and is defined as:
\begin{equation}
\small
\begin{aligned}
     \text{Performance} (\%) \times \text{Quality} (\%) \times \text{Availability} (\%) = \text{OEE} (\%).
\end{aligned}
\end{equation}
We assume an \emph{availability}, which is the ratio between actual and planned production time, of 100\%, due to limited access to the factors affecting this index. The \emph{performance} and \emph{quality} are defined as:
\begin{equation}
\label{eq:OEE performance}
\small
\begin{aligned}
     \text{Performance} =  P_{p} 
      / P_{s} \times 100\%, 
\end{aligned}
\end{equation}
\begin{equation}
\small
\begin{aligned}
\label{eq:OEE quality}
     \text{Quality} =  B_{p} 
      / B_{s}  \times 100\%. 
\end{aligned}
\end{equation}
The term $P_{p}$, \emph{packed products}, denotes the number of products that exit the robotic packaging machine enclosed in a box, and  $B_{p}$, \emph{packed boxes}, refers to the boxes that leave the machine containing the required number of products. Similarly, terms $P_{s}$ and $B_{s}$ denote products and boxes that enter the machine.

\subsection{System Overview}
Fig. \ref{fig:machineoverview} depicts a schematic overview of our specific configuration. It contains four delta robots (denoted by R1-R4), and their workspaces depicted by the rectangles below them.
The products and boxes enter the machine at the box and product detection point and leave the machine at the checkout points, respectively.
Both conveyor belts move in the same direction, with a constant speed of $v_P$ for the conveyor belt with two product lanes and a controllable speed of $v_B$  for the conveyor belt with boxes.
Two delta robots are used for filling each box, where each robot is responsible for one of the two layers in that box. 
The control system of the packaging machine assigns the robots, with the products in the horizon and the boxes on the conveyor belt.

\subsection{Belt Speed Optimization Problem}

\begin{figure}[t]
\centerline{\includegraphics[width=1.0\linewidth]{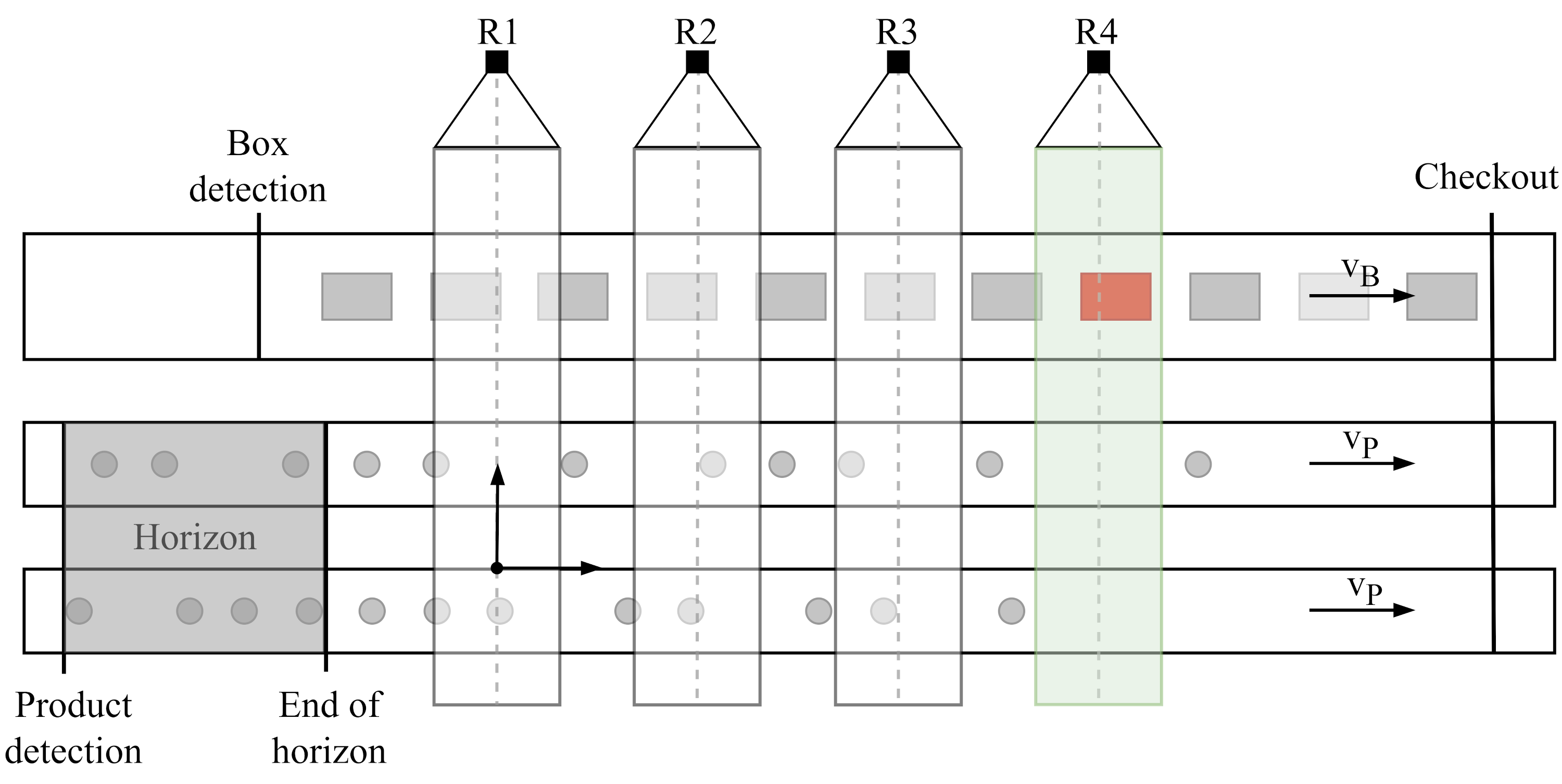}}
\caption{Schematic view of the robotic packaging machine with one box belt at the top and one product belt with two lanes at the bottom.}
\label{fig:machineoverview}
\end{figure}

For a constant product inflow and constant product and box belt speeds, the machine is able to comply with the requirements stated in Subsection \ref{subsec: robotic packaging machine}. However, in the case of varying product inflow and constant belt speeds, the delta robots may not be able to reach the necessary products when an empty or partially filled box is about to leave the machine, which results in a violation of the requirements. This issue is illustrated by the red box in Fig. \ref{fig:machineoverview} and can lead to waste and critical problems later in the packaging process, such as during box sealing.

To ensure compliance with the requirements even under varying product inflow, the box belt speed must be optimized according to the changing product inflow. To achieve this, we propose the following constrained minimization problem:
\begin{subequations}
\begin{flalign}
    \min _{v_B[k]} \sum_{k=1}^{N}P_{l}[v_B[k], k]  \label{subeq:opti1} & \\
    \text { s.t. } \sum_{k=1}^{N}P_{l}[v_B[k], k] / \sum_{k=1}^{N}P_{s}[k]  \leq & 1- 0.998 &\label{subeq:opti2}\\
    \sqrt{(v_B[k]-v_B[k-1])^2} / \Delta k \leq & \enskip a_{B,\mathrm{max}} &\label{subeq:opti3}\\
    B_{le}\left[v_B[k], k\right]  = & \enskip0 & \label{subeq:opti4}\\
    B_{lp}\left[v_B[k], k\right]  = &\enskip 0 & \label{subeq:opti5}\\
    P_{s}[k] > &\enskip 0 & \label{subeq:opti6}\\
    v_B[k]  \in & [v_{B,\mathrm{min}},v_{B,\mathrm{max}}] \label{subeq:opti7}& 
\end{flalign}
\label{eq:constrained criterion}
\end{subequations}
where $P_l$ represents the number of lost products, 
$B_{le}$ lost empty boxes, and $B_{lp}$ lost partly filled boxes
for each time step $k$. We define as lost a product, empty box or partly filled box when it has passed the checkout point, which is denoted in Fig. \ref{fig:machineoverview}. The objective function \eqref{subeq:opti1} minimizes the number of lost products during an operating time of $N$ seconds of the machine by optimizing the box belt speed at each time step. By definition, this results in a maximization of placed products per minute, because the product belt speed, $v_P$, nor the product inflow can be controlled. The first requirement stated in Section \ref{subsec: robotic packaging machine} is represented by constraint \eqref{subeq:opti2}, which limits the maximum allowable percentage of product loss during machine operation of $N$ seconds to be lower than 0.2$\%$. The two requirements regarding the boxes are represented by constraint \eqref{subeq:opti4} and \eqref{subeq:opti5}. An additional constraint is necessary to limit the acceleration of the box belt, which is denoted by \eqref{subeq:opti3} and calculates the Euclidean distance between two successive speeds and divides this by the time step length $\Delta k$. The maximum box belt acceleration is denoted by $a_{B,\mathrm{max}}$. Solving  \eqref{eq:constrained criterion} yields the box belt speed profile, $v_B[k]$, for an operating time of $N$ seconds of the robotic packaging machine, maximizing the number of placed products above 99.8\% while ensuring that no empty or partly filled boxes are lost and the box belt acceleration remains in the feasible range.

To solve the constrained minimization problem \eqref{eq:constrained criterion} with RL, we re-formulate the problem as a Markov Decision Process (MDP) with penalty functions, with discrete time steps, $k=0,1, \ldots \in \mathbb{Z}$ and time step length $\Delta k$. The penalty functions encourage constraint satisfaction \cite{b4} \cite{b53}. Since RL problems are modelled as a maximization problem, the employed reward is denoted as $-P_l[v_B[k], k]$ plus the penalty functions. The objective of the learning agent is to obtain the optimal control policy $\pi^*$, which is a mapping from states $s_k$ to actions $a_k$, by maximizing the cumulative reward.

%
%
\section{Methodology} 
\label{sec:methodology}
In order to solve this problem with RL, we need to make multiple important design decisions. To successfully learn an RL policy, that works well on real packaging machines, we introduce a framework that takes into account penalty functions, control delays, observation matching, state representation design, sparse and delayed rewards and smooth control. In the following subsections, we provide details regarding our approach to addressing each one of these challenges. 

\subsection{State and Action Representation Design} 
\label{subsec:markovstate}
After extensive experimentation, we settle on this action representation.
The action space, denoted by $A$, is continuous, normalized, and symmetric, and is used to represent the box belt speed, $v_B$. To ensure that the machine's allowed control inputs are adhered to, we subsequently employ one-dimensional linear interpolation to rescale the actions to fall within the range of $v_{B,\mathrm{min}}$ and $v_{B,\mathrm{max}}$.

Furthermore,  we conducted experiments to select features and history for designing a state that captures all relevant information for selecting appropriate actions. The final features are:

\begin{itemize}
    \item Current box belt speed, $v_B[k]$ [m/s],
    \item Previous box belt speed, $v_B[k-1]$ [m/s],
    \item Product inflow lane 1 measured at product detection [products/min],
    \item Product inflow lane 2 measured at product detection [products/min],
    \item Distance of closest unassigned empty box relative to the checkout point, $x_\mathrm{box}$ [m],
    \item Distance of closest unassigned product relative to checkout point, $x_\mathrm{prod1}$ [m],
    \item Distance of second closest unassigned product relative to checkout point, $x_\mathrm{prod2}$ [m].
\end{itemize}
Where each feature is represented by a continuous normalized and symmetric feature space.

To handle partial observability, introduced by the non-stationarity and stochasticity of the system, 30 time steps of history are added to each feature, which captures a complete throughput of products from detection to checkout. To tackle the increased complexity,  resulting from the high-dimensional continuous state space, neural networks are used for function approximation in the RL agent.

\subsection{Reward Design}
\label{subsec:sparsereward}

Two simplifications and one potential problem can already be uncovered for the constrained optimization problem \eqref{eq:constrained criterion}. 
First, we simplify the optimization problem by removing constraint (\ref{subeq:opti2}) regarding the maximum number of lost products. This constraint is cumulative and forms the lower bound of the objective function (\ref{subeq:opti1}). Therefore, we state that if a global optimum exists for the optimization problem (\ref{eq:constrained criterion}), it is equal to the global optimum of the same optimization problem without this cumulative constraint. Therefore, we can neglect this constraint, which also makes constraint (\ref{subeq:opti6}) redundant.  
Second, we eliminate the constraint on partly filled boxes \eqref{subeq:opti5}. Specifically, the packaging machine prioritizes the delivery of fully filled boxes, thereby precluding the intentional departure of partly filled boxes from the machine. However, if our RL agent interferes with the internal control scheme, variations in the box belt speed during pick and place task executions could lead to the unintentional departure of partly filled boxes. 

Since the agent's actions are continuous, it has the ability to control the box belt speed with high precision. However, this could lead to the emergence of multiple (weak) global optima and an increase in the number of weak local optima; small speed changes that do not impact the return, but do negatively affect the box belt's maintenance. While smooth control is not a strict requirement, it is preferable for stable training, real-world applicability, and enhanced interpretability of the resulting policy \cite{b8}. Therefore, we remove inequality constraint \eqref{subeq:opti3} from the constrained optimization problem \eqref{eq:constrained criterion} and add penalty function \eqref{eq:penaltyacc} that penalizes speed changes with an appropriately small amount $\zeta$, this way smooth control is encouraged. 
\begin{equation}
\begin{aligned}
p(v_B)&=-\zeta \,\sqrt{(v_B[k]-v_B[k-1])^2} \\ 
\end{aligned}
\label{eq:penaltyacc}
\end{equation}
By incorporating the simplifications and the penalty function for smooth control, the combined reward function becomes:
\begin{equation}
\begin{aligned}
    r =   &- \mu_\mathrm{prod}  \cdot P_l[v_B[k], k] - \mu_\mathrm{box}  \cdot B_{le} [v_B[k], k], \\
    & + p(v_B). 
\end{aligned}
\label{eq:rewardfunctioncase2}
\end{equation}
where $\mu_\mathrm{prod}$ denotes the weight for the number of lost products \eqref{subeq:opti1}, $\mu_\mathrm{box}$ denotes the weight for the constraint on empty boxes \eqref{subeq:opti4} and $p(v_B)$ being defined in \eqref{eq:penaltyacc}.

\subsection{Action Delay and Observation Matching}  
\label{subsec:controldelay}
System delays are fundamental challenges of real-world RL. They negatively influence both the learning process and final solution \cite{b9}, \cite{b7}. 
Delays in RL can be divided into three categories: action, observation and reward delays \cite{b8}. 

\begin{figure}[t]
\centering
    \centerline{\includegraphics[width=1.0\linewidth]{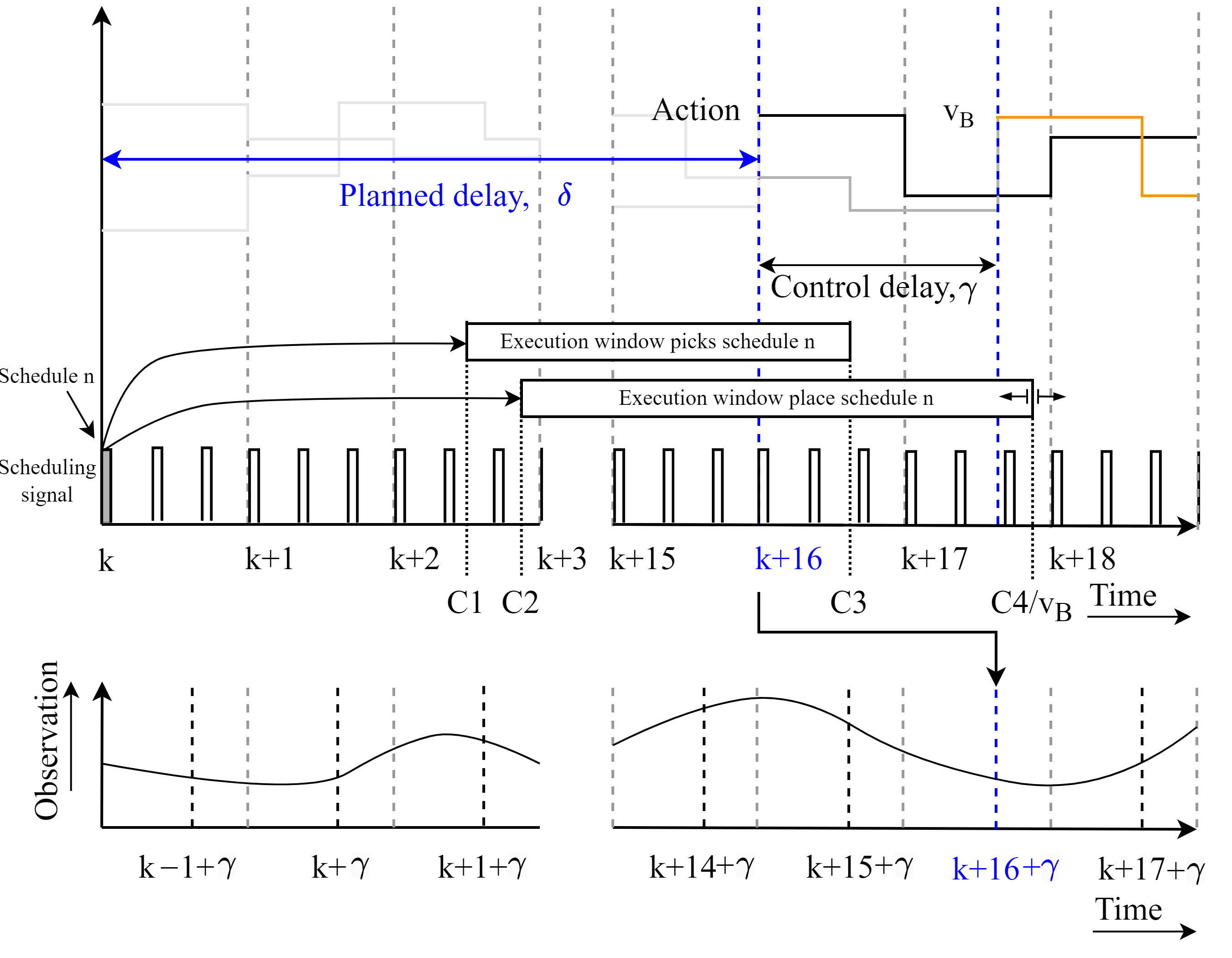}}
    \caption{Overview of the control delay, planned delay and observation matching.}
    \label{fig:Time_horizon_all}
\end{figure}

In this work, we mainly deal with action delays. We distinguish two types of action delays: \emph{control delay} and \emph{planned delay}. Control delay $\gamma$ is inevitable, due to the machine's internal dynamics, and represents the time interval between action selection by the RL agent and the actual execution of that action by the packaging machine. In contrast to control delay, we intentionally introduce planned delay $\delta$, in order to minimize the interference between RL control loop and the machine's control scheme. 
To illustrate the problem of interference, we need to describe internal machine assignment dynamics.
Fig. \ref{fig:Time_horizon_all} shows the action signal, box belt speed $v_B$, and the control delay $\gamma$ and planned delay $\delta$. 
Moreover, the scheduling signal of the packaging machine is displayed. 
A schedule $n$ is related to \emph{pick and place execution windows}. The pick execution window is defined as the time interval between the moment a product enters the workspace of the first robot, denoted as $C1$, and the moment a product leaves the workspace of the last robot, denoted as $C3$. Additionally, the place execution window is defined as the time interval between the earliest possible time the first products of schedule $n$ can be placed into a box, denoted as $C2$, and the latest possible time the last products of schedule $n$ can be placed in a box, denoted as $C4/v_B$. $C1$, $C2$ and $C3$ are constant (future) time instances, relative to the creation time of the corresponding schedule $n$, because they only depend on the constant product belt speed $v_P$ and the fixed configuration of the machine. $C4$ denotes the latest possible box position for which the last products of schedule $n$ can be placed. To get the respective time, $C4$ is divided by the variable box belt speed $v_B$.

In our initial experiments, we could observe that the RL agent learns to apply only minimum actions (keeps constant belt speed) in order to avoid negative interference with the rest of the control system.
Therefore, we intentionally add the \textit{planned delay} $\delta$ to the control delay such that the execution of the action takes only place after the execution window of all picks of schedule $n$. In this way, the speed profile of the box belt is known to the machine during the job assignment and the RL agent does not interfere with the control system. As can be seen in Fig. \ref{fig:Time_horizon_all}, the placement of products in a box can still happen after the combined control and planned delay; however, this will only happen for low product inflow. As stated earlier, the length of the execution window for a place is not fixed, but dependent on the box belt speed $v_B$. For low product inflow, the box belt speed must be low in order to minimize the number of unfilled boxes, which increases the length of the place execution window. In contrast, the constant product belt speed is relatively high, therefore, the delta robots must hold the products for a while before placing them in a box. Hence, the pick part of the schedules remains valid and the place part of the schedule is updated according to the changing box belt speed until a box enters the workspace of the corresponding delta robot.
%
%

The delays introduce a time mismatch between action execution, received observation and rewards. To simplify the learning problem, the observation at time $ k +\delta + \gamma$ is fed back to the agent at time $k$, which is shown in Fig. \ref{fig:Time_horizon_all}. This future observation matching method is possible because we have available future assignments of all the products in the horizon; therefore, all the information of a schedule created at the time $k$, (i.e., the future start and end times), can be used to construct the observation at time $ k +\delta + \gamma$. 
This approach is similar to the concept of Smith predictor \cite{b99}.

\section{Experimental Results}
In this section, we describe the experimental setup and present the experimental validation of the learned policy, computed with our proposed method, on real-world product inflow data and compare its results with a rule-based engineered baseline solution.

\subsection{Experimental Setup} 
The learning process is carried out using simulation. Nevertheless, the scenarios encountered in the simulation must be realistic so that the learned policy can transfer properly to the real world. To address this, we employ available information about the machines and their usage to train our agent in a wide range of scenarios that are encountered in reality. More specifically, secondary packaging solutions mostly operate under product inflows per lane between 120 and 135 products per minute. Therefore, in each training episode, we apply \emph{scenario randomization} by randomizing the product inflow in this range. To ensure the effectiveness of our approach, we validate our method using real-world data from over seven different challenging scenarios. This data comprises information such as product and box detection times, as well as the start and end times of schedules. Note that this data is not enough to learn from it using approaches such as offline RL \cite{b16}, since it amounts to 75.5 minutes of operational time. However, it is enough to validate our RL agent on the scenarios found in reality.

As the RL algorithm, we choose Proximal Policy Optimization (PPO) due to its combination of sample efficiency and performance \cite{b17}. Furthermore, inspired by the work of \cite{b18}, \cite{b49} and \cite{b50} we use a combination of the adaptive learning rate optimizer Adam and a linear decaying learning rate schedule. The RL policy is trained in 6827 episodes of 1800 simulated seconds per episode, equivalent to approximately 142 days of machine operation. The training time is approximately nine hours using 16 parallelized simulations on an Intel Core 11th Generation i7-11850H processor with 32GB RAM. Additionally, we conducted a hyperparameter study to optimize training performance. 
%
 
\subsection{Validation} 
We validate the learned policy in realistic scenarios using real-world product inflow data and compare it with the rule-based engineered solution. The rule-based solution has been employed for several years, making it a suitable benchmark for this research. Due to the limited availability of real-world product inflow data, we train the RL solution using simulated randomized scenarios, as discussed earlier and validate its performance in realistic scenarios with real-world data.

Table \ref{tab:rwd} presents several metrics for comparing the RL solution with the \emph{rule-based engineered solution}. The metrics include two indexes of the OEE industry standard metric, namely performance and quality. Performance measures the quantity of successfully processed products, while quality represents the quality of the outgoing boxes in terms of filling rate. 


\renewcommand{\arraystretch}{1.0}
\begin{table}[hbtp]
\fontsize{6.3}{8}\selectfont
\caption{Validation of two policies trained with reinforcement learning and the rule-based engineered baseline solution using real-world inflow data}
\begin{center}
\begin{tabular}{|c|c|c|c|c|}
\hline
&\multirow{3}{*}{\textbf{Metrics:}}&\multicolumn{2}{|c|}{\textbf{Method}} & \multirow{2}{*}{\text{RL}}\\
\cline{3-4} 
&\multirow{3}{*}{\text{Mean (Std.)}}& \multicolumn{1}{|c|}{\text{1: Reinforcement}} & \multicolumn{1}{|c|}{\text{2: Rule-Based}}& \multirow{2}{*}{\text{vs.}} \\
& &\text{Learning (ours)} & \text{Engineered}& \multirow{2}{*}{\text{Baseline}} \\
&&\text{}& \text{Baseline} &\\
\hline

Productivity & \multirow{1}{*}{\text{Performance$^{\mathrm{a}}$ [\%] }} & \multirow{1}{*}{99.94  (0.077)} & \multirow{1}{*}{99.31  (0.22)} & \multirow{1}{*}{+0.63\%}\\
&\multirow{1}{*}{\text{Quality$^{\mathrm{b}}$ [\%]}}  & \multirow{1}{*}{100.00  (0.00)}&  \multirow{1}{*}{100.00 (0.00)}& \multirow{1}{*}{-} \\
&\multirow{1}{*}{\text{Lost products [prod]}}  & \multirow{1}{*}{1.28 (1.58)}&  \multirow{1}{*}{ 19.00 (15.64)}& \multirow{1}{*}{-93.26\%} \\

\hline



Control input & \multirow{2}{*}{\text{Mean $a_{B}^{\mathrm{c}}$ [$m/s^2$]}}& \multirow{1}{*}{3.21$\cdot10^{-3}$} & \multirow{1}{*}{18.55$\cdot10^{-3}$}& \multirow{2}{*}{-82.70\%}\\
& & \multirow{1}{*}{(0.25$\cdot10^{-3}$)} & \multirow{1}{*}{(1.07$\cdot10^{-3}$)}&\\

&\text{Computation$^{\mathrm{d}}$} &  \multirow{2}{*}{9.92 (0.49)} &  \multirow{2}{*}{18.02 (0.60)}& \multirow{2}{*}{-55.05\%} \\
&\text{time [ms/s]} &&& \\
\hline
Constraints$^{\mathrm{e}}$ &\fontsize{5.8}{8}\selectfont\text{Performance} $\geq$ 99.8\% & \checkmark & $\times$&\\

&\fontsize{5.8}{8}\selectfont\text{Empty boxes = 0} & \checkmark & \checkmark &\\

&\fontsize{5.8}{8}\selectfont\text{Partly filled boxes = 0}& \checkmark & \checkmark &\\

&\text{$a_B^{\mathrm{c}}\leq a_{B, \max }$} & \checkmark & $\times$& \\

\hline
\multicolumn{5}{l}{$^{\mathrm{a}}$(Number of packed products / total supplied products)*100\%.}\\
\multicolumn{5}{l}{$^{\mathrm{b}}$(Number of packed boxes / total supplied boxes)*100\%}\\
\multicolumn{5}{l}{$^{\mathrm{c}}$ Box belt acceleration}\\
\multicolumn{5}{l}{$^{\mathrm{d}}$ Computation time between start and end of single simulation [ms]} \\
\multicolumn{5}{l}{\quad  / simulated operating time of machine [s]. } \\
\multicolumn{5}{l}{$^{\mathrm{e}}$ If zero constraint violation is obtained for each of the seven simulations.} \\
\end{tabular}
\label{tab:rwd}
\end{center}
\end{table}

Table \ref{tab:rwd} shows that the RL solution outperforms the baseline solution in terms of performance, mean box belt acceleration, and computation time. Specifically, the RL solution increases performance by $0.63\%$, resulting in a $93.26\%$ decrease in lost products, decreases mean box belt acceleration by $82.70\%$, and decreases computation time by $55.05\%$, while having zero constraint violations. In contrast, the baseline solution violates performance and maximum box belt acceleration constraints in every simulation with real-world scenarios.

\begin{figure}[htpb]
\centering
\begin{subfigure}{0.5\textwidth}
    \includegraphics[width=0.957641\linewidth]{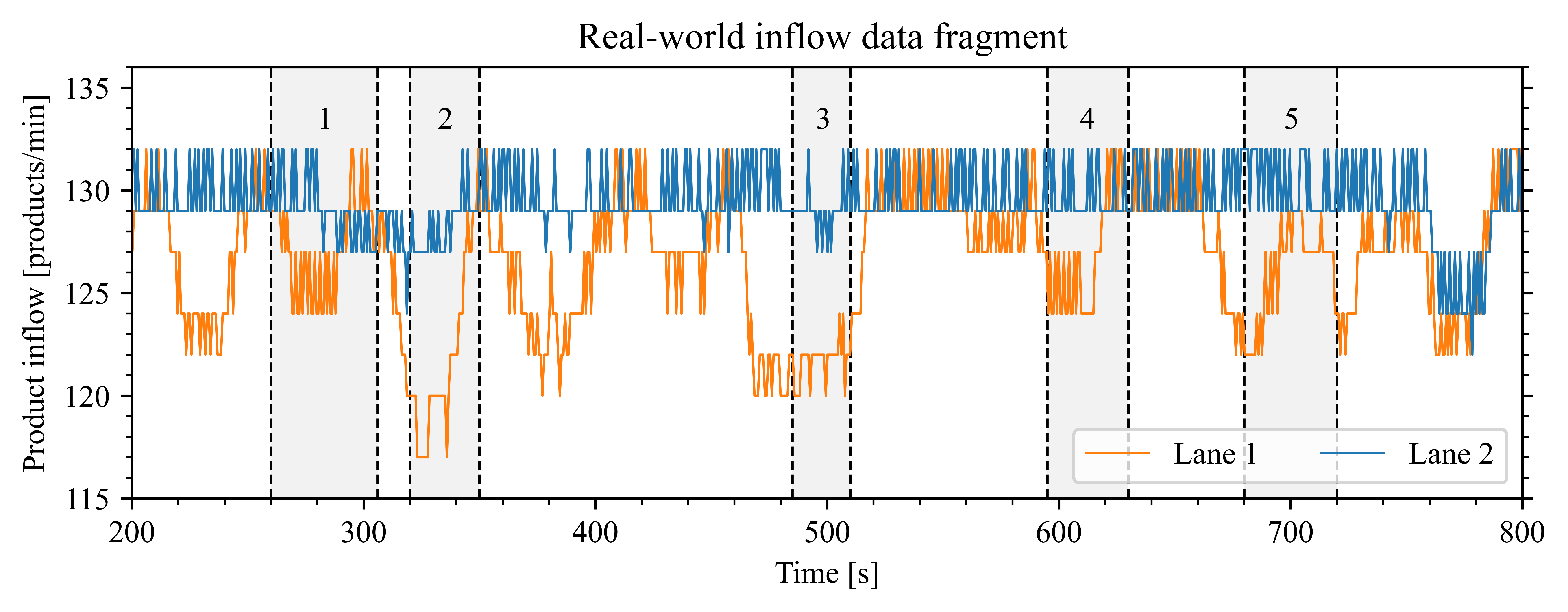}
    \label{fig:Inflow}
\end{subfigure} 
\hfill
\begin{subfigure}{0.5\textwidth}
    \centerline{\includegraphics[width=1.0\linewidth]{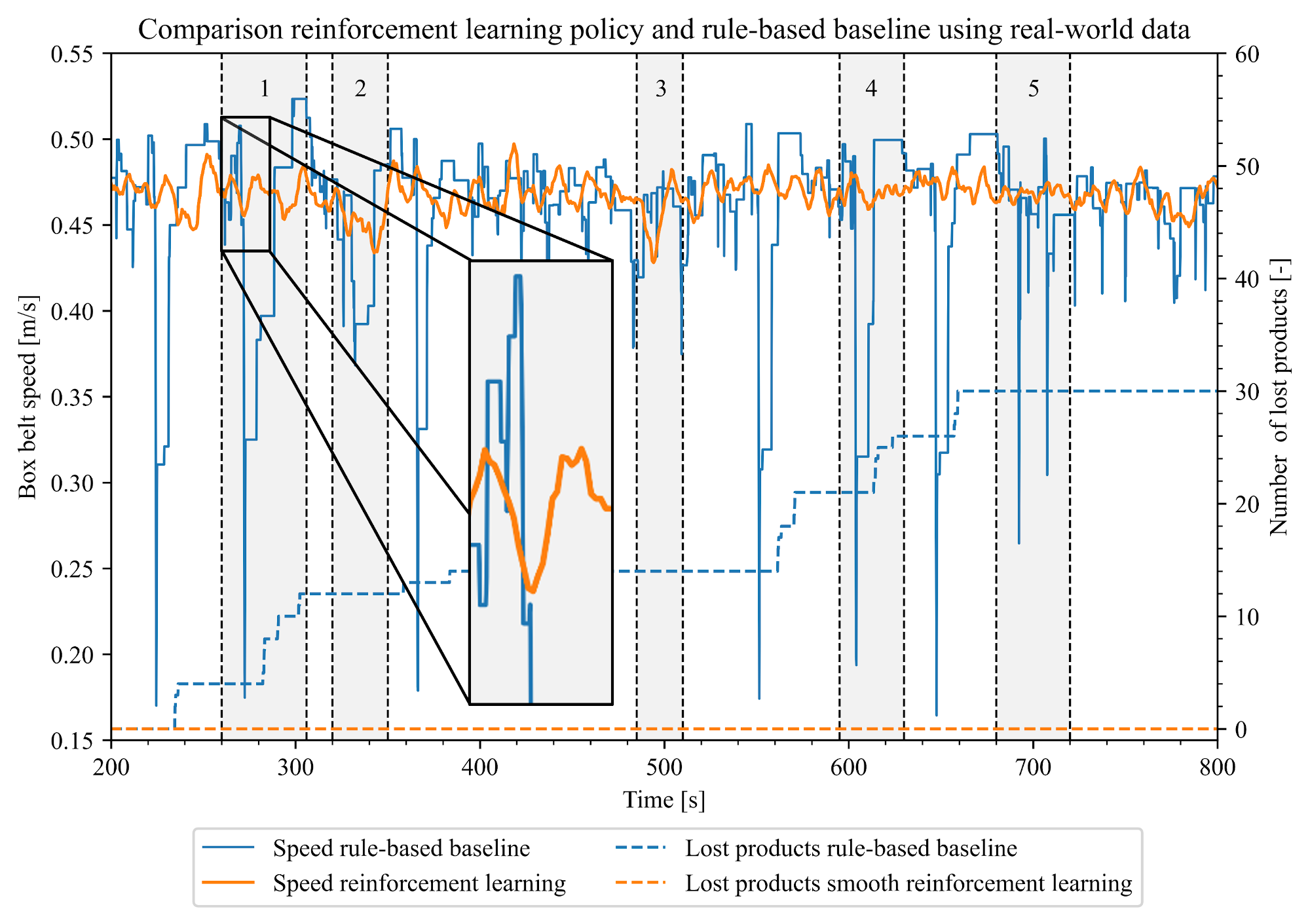}}
    \label{fig:SpeedCase3}
\end{subfigure}
\caption{Fragment of real-world inflow validation data with corresponding speed plots and the number of lost products for the cases listed in Table \ref{tab:rwd}. }
\label{fig:speedplotsRWD}
\end{figure}

Fig. \ref{fig:speedplotsRWD} demonstrates the differences between the RL solution (orange line) and the baseline solution (blue line) in one scenario. The product inflow rate is shown at the top of the figure. It can be seen that the baseline solution exhibits corrective behavior, adjusting box belt speed only when critical situations occur, eventually resulting in product loss. This is evident in regions 1, 2, 4, and 5, where the box belt speed decreases drastically to prevent boxes from leaving partly or unfilled, causing increased product loss (dashed line regions 1 and 4). In contrast, the RL solution shows a smaller and smoother response to variations in product inflow. It is able to prevent drastic slow-down by slightly adapting the speed in a predictive fashion, based on the product inflow and relative box and product locations on the conveyor belts, before the critical situations even occur. An instance of the proposed preventive strategy is demonstrated in the close-up of region 1. As can be seen, the baseline method exhibits myopic behavior by first increasing its speed, leading to a drastic speed drop a bit later, ultimately resulting in product loss (blue dashed line region 1). Thus, the RL solution completes the scenario with zero product loss compared to the 30 products lost with the baseline solution.
Furthermore, the RL solution provides speed changes with a smaller amplitude and lower frequency, resulting in smoother control (region 5). 

Additionally, we expected the policy to oscillate to some degree because lower product inflows require also lower box inflow, leading to lower box belt speeds, and vice versa. As seen from the product inflow data presented in Fig. \ref{fig:speedplotsRWD}, there are oscillations to some extent, and this is indeed reflected in the RL policy. Specifically, we observe that the decrease in box belt speed matches the decrease in product inflow, marked as regions 1, 2 and 3 in Fig. \ref{fig:speedplotsRWD}. Notably, the learned policy generalizes well to out-of-distribution product inflows, since it was trained for inflows between 120 and 135 products per minute per lane, and the policy properly addresses edge-case scenarios where the inflow drops to 115.

In addition, the mean computation time is decreased by $55.05\%$ compared to the baseline solution, as demonstrated in Table \ref{tab:rwd}. It is also notable that the standard deviation of the computation time is significantly lower for the RL solution compared to the baseline solution, as it computes the policy more consistently.

Thus, the validation study conducted using real-world product inflow data shows that our proposed framework effectively learns appropriate box belt speed policy for a simulated robotic packaging machine in realistic scenarios. This is achieved without any violation of performance, quality, or acceleration constraints, as evident from Method 1 in Table \ref{tab:rwd}. Furthermore, we can state that the proposed scenario randomization method generalizes well to realistic scenarios. 

\section{Conclusions} 
In this work, we investigate the feasibility of RL in the robotic industry by solving a complex robotic optimization problem from the food packaging industry. We propose a method that uses a planned delay to enable the integration of RL with minimal interference in complex control schemes. We demonstrated that the proposed framework achieves zero constraint violations in all simulations based on real-world product inflow data. 

Contrary to the rule-based  baseline, the proposed framework complies with all the given requirements set by the industry. In the validation study, performance surpasses the required $99.8\%$, the quality maintains at $100\%$ and the accelerations are kept in the feasible range. This means that $99.8\%$ of the supplied products are packed in a box, and no boxes leave the machine empty or partially filled. Compared to the rule-based baseline, our proposed solution improves performance by $0.63\%$, resulting in a decrease in product loss of $93.26\%$, while also decreasing the mean acceleration and computation time by $82.70\%$ and $55.05\%$, respectively. 

We conclude that the framework is able to deal with control delays, sparse delayed rewards and policy inference in the complex interdependent control scheme of the packaging machine. With the proposed method, predictive behavior for varying product supply is learned while satisfying the machine's performance and quality constraints. The design encourages smooth control of the conveyor belt, reducing maintenance and increasing the interpretability of learned policy. Additionally, it requires fewer time-demanding decisions during the operating time of the machine. Moreover, it handles the limited availability of real-world product inflow data well by using solely simulated product inflow data and scenario randomization for training.

Future work should investigate how the learned policy computed with the proposed framework can be transferred to a physical robotic packaging machine. While this research has partially bridged this gap by using scenario randomization and real-world product inflow data, further efforts are necessary to address the remaining aspects of this gap. For instance, it would be ideal to have more real-world product inflow data to use a combination of scenario randomization and real-world data in the training process, as done in \cite{b33}. Novel situations that are hard to capture with scenario randomization solely for reasonable training times, such as product inflow stops or machine startup issues, are then added to the training data set. On the other hand, collecting enough real data from the machine takes time.

\section*{Acknowledgment}
This research is conducted in collaboration with BluePrint Automation. They provided access to the packaging machine's simulator, data and practical information regarding the food packaging industry. Their cooperation is hereby gratefully acknowledged. This research is partially funded by the Netherlands Organization for Scientific Research project Cognitive Robots for Flexible Agro-Food Technology, grant P17-01 and the European Research Council Starting Grant TERI, project reference \#804907.

\bibliographystyle{IEEEtran}
\bibliography{case}

\end{document}